\documentclass{bmvc2k}

\title{Learning Part Motion of Articulated Objects Using Spatially Continuous Neural Implicit Representations}

\addauthor{Yushi Du*}{1900012147@pku.edu.cn}{1}
\addauthor{Ruihai Wu*}{wuruihai@pku.edu.cn}{1}
\addauthor{Yan Shen}{yan790@pku.edu.cn}{2}
\addauthor{Hao Dong\dag}{hao.dong@pku.edu.cn}{3}

\usepackage{graphicx}
\usepackage{amsmath}
\usepackage{amssymb}
\usepackage{booktabs}
\usepackage{wrapfig}
\usepackage{graphicx,caption}
\usepackage{array}
\usepackage{tabularx}

\addinstitution{
 Center on Frontiers of Computing Studies\\
 School of Computer Science\\
 Peking University\\
 \ \\
 * \ denotes equal contribution \\
 \dag \ denotes corresponding author
}

\runninghead{Yushi Du*, Ruihai Wu*, Yan Shen, Hao Dong\dag}{Learning Part Motion}

\begin{document}

\maketitle

\begin{abstract}
Articulated objects (\emph{e.g.}, doors and drawers) exist everywhere in our life. 
Different from rigid objects, articulated objects have higher degrees of freedom and are rich in geometries, semantics, and part functions.
Modeling different kinds of parts and articulations with nerual networks plays an essential role in articulated object understanding and manipulation, and will further benefit 3D vision and robotics communities.
To model articulated objects, most previous works directly encode articulated objects into feature representations, without specific designs for parts, articulations and part motions.
In this paper, we introduce a novel framework that explicitly disentangles the part motion of articulated objects by predicting the transformation matrix of points on the part surface, using spatially continuous neural implicit representations to model the part motion smoothly in the space. 
More importantly, while many methods could only model a certain kind of joint motion (such as the revolution in the clockwise order),
our proposed framework is generic to different kinds of joint motions in that transformation matrix can model diverse kinds of joint motions in the space.
Quantitative and qualitative results of experiments over diverse categories of articulated objects demonstrate the effectiveness of our proposed framework.
\end{abstract}


\vspace{3mm}
\section{Introduction}
\label{sec:intro}

There are a plethora of 3D objects around us in the real world.
Compared to those rigid objects with only 6 degrees of freedom (DoF), articulated objects (\emph{e.g.}, doors and drawers) additionally contain semantically and functionally important articulated parts (\emph{e.g.}, the screen of laptops), 
resulting in their higher DoFs in state space, and more complicated geometries and functions. 
Therefore, understanding and representing articulated objects with diverse geometries and functions is an essential but challenging task for 3D computer vision.

Many studies have been investigating the perception of 3D articulated objects, including discovering articulated parts~\cite{gadre2021act, huang2021multibodysync, nie2022structure}, inferring kinematic models~\cite{abbatematteo2019learning, katz2013interactive}, estimating joint configurations~\cite{jain2021screwnet, liu2020nothing, li2020category, desingh2019factored}, predicting part poses~\cite{yan2020rpm, wang2019shape2motion}, building digital twins~\cite{jiang2022ditto} and manipulating parts~\cite{xu2022umpnet}. 
One recent work, A-SDF~\cite{mu2021sdf}, studies the representations of articulated objects by encoding shape and articulation into latent space. But instead of considering modeling articulation objects as linking parts under motion constraints, they directly decode the whole object point cloud into the latent space.
%
Another work, Ditto~\cite{jiang2022ditto}, successfully generates objects under novel poses over diverse joint motions (\emph{e.g.}, rotation and displacement over different axis) using a single network. 
However, this method relies on specific articulation annotations such as joint type, orientation, and displacement which limits their ability to generalise across diverse articulations (\emph{e.g.}, different joint motion and type).


\begin{figure}
\vspace{-1mm}
\begin{tabular}{ccc}
\includegraphics[trim={180, 220, 200, 140}, clip, scale=0.33]{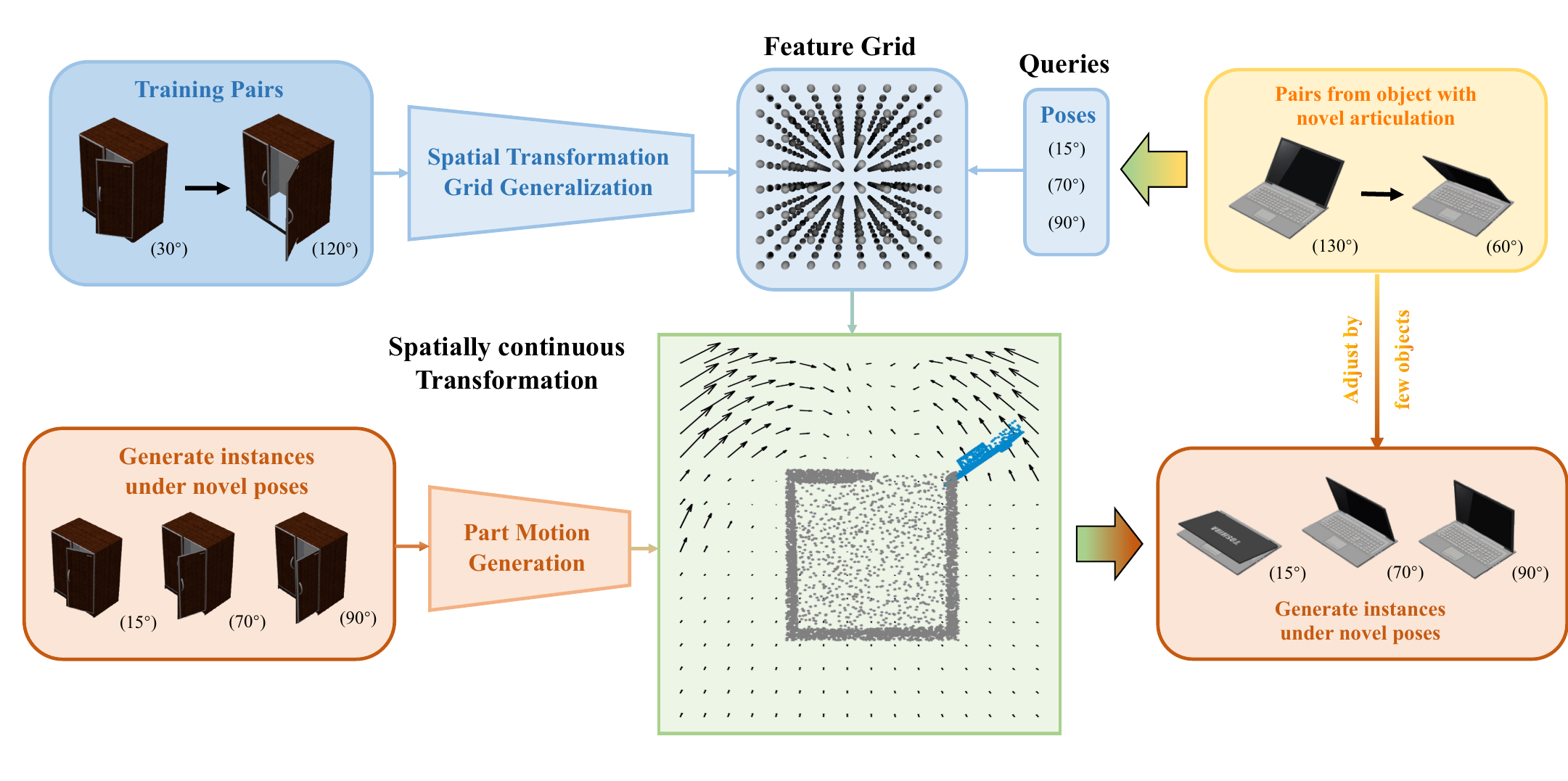}
\end{tabular}
\caption{We propose \textbf{spatially continuous neural implicit grid} that receives two point clouds of the same object under different part poses. 
The point clouds are provided with their corresponding articulated part poses and the grid could encode two frames of point clouds into a spatially continuous implicit feature grid with both geometric and pose information. By taking different new part poses as queries, we decode per-point transformations representing articulated part motions from the feature grid. Then we move the object using the transformation to generate objects under novel poses.
This representation could be easily adjusted to articulated objects with novel shapes and joint motions (\emph{e.g.}, from door to laptop) tuned on only a few new objects.}
\vspace{-0.6cm}
\label{fig:teaser}
\end{figure}


In this paper, we introduce a novel framework for learning a spatial continuous representation of the part motion of articulated objects, and enable the few-shot generalisation across different novel object categories with different joint motion. 
To be specific, we model articulation as a constraint that can map a scalar value representing the part poses to a transformation describing the movements of the articulated parts.

To further study the representations of articulated objects, with a focus on the objects' parts, we introduce our novel framework for learning the part motions of articulated objects. To be specific, we model the movement of parts as a mapping between a scalar representing the part pose and a transformation matrix. 
For a reason that part motion is a core and generic property shared by all articulated objects, our proposed framework is generic to various articulated objects with diverse kinds of part motions, without any need to have specific designs for each kind of object.

Considering the limited number of DoF of joints on articulated objects, the motions of points on the articulated part should make up a continuous and smooth distribution with respect to points' positions on parts. In other words, close points on the part surface have similar motions, while far away points have varied motions.
Therefore, we further propose to use spatially continuous neural implicit representations for the representations of point motions on the articulated part.
Inspired by ConvONet~\cite{peng2020convolutional}, we build a fine-grained and spatially continuous implicit grid for learning the representations of point-level transformations from one pose to another.

We conduct experiments over large-scale PartNet-Mobility dataset~\cite{chang2015shapenet, Mo_2019_CVPR, Xiang_2020_SAPIEN}, covering 3D articulated objects with diverse geometries over 7 object categories. 
Quantitative and qualitative results demonstrate that using the spatially continuous grid, our method accurately and smoothly models part motion and generates articulated objects with novel part poses reserving detailed geometries, showing our superiority over baseline methods. 
%

\vspace{-1mm}
\section{Related Work}
\label{sec:related_work}
\vspace{-1mm}

\subsection{Representing Articulated Objects}
How to understand and to model articulated objects has been a long-lasting research topic, including segmenting articulated parts~\cite{katz2013interactive, huang2012occlusion, martin2014online, katz2008manipulating}, tracking feature trajectories~\cite{katz2013interactive, hausman2015active, desingh2019factored}, estimating joint configurations~\cite{martin2014online, katz2014interactive, martin2016integrated, hausman2015active, katz2008manipulating}, and modelling kinematic structures~\cite{martin2014online, martin2016integrated, sturm2011probabilistic, michel2015pose}.
Recently, many works~\cite{li2020category, huang2021multibodysync, abbatematteo2019learning, jain2021screwnet, liu2020nothing, yan2020rpm, wang2019shape2motion, zeng2021visual} further utilise the deep learning methods to study diverse articulated objects, leading to better performance and stronger generalisation.
A recent work A-SDF~\cite{mu2021sdf} studies the problem of generic articulated object synthesis and leverages implicit functions to decode articulated objects into latent codes. 
However, most of these works represent articulated objects by abstracting standardised kinematic structure, estimating joint parameters, and predicting part pose, which may not provide explicit information on articulated shapes for downstream tasks like robotics manipulation~\cite{gadre2021act, mo2021where2act, wu2022vatmart, wang2022adaafford, eisner2022flowbot3d, wu2023learning, ning2023where2explore, geng2022gapartnet, geng2023partmanip}. 
Different from those works, we utilise neural implicit functions for explicit articulated object representation and generation.

\vspace{-1mm}
\subsection{Neural Implicit Representation} 

A vast and impressive literature has investigated neural implicit representations~\cite{chen2019learning, genova2020local, jiang2020local, mescheder2019occupancy, mildenhall2021nerf, niemeyer2020differentiable, park2019deepsdf, sitzmann2019scene, peng2020convolutional}, which utilises deep neural networks to implicitly encode 3D shapes into continuous and differential signals in high resolution. 
While most of the previous works study the representation of 3D rigid objects, two recent works, A-SDF~\cite{mu2021sdf} and Ditto~\cite{jiang2022ditto}, focus on the representation of 3D articulated objects. A-SDF~\cite{mu2021sdf} represents the articulated objects by separately encoding shape and articulation into latent space. Ditto~\cite{jiang2022ditto} builds digital twins of articulated objects by reconstructing the part-level geometry and estimating the articulations explicitly. 
However, both of the works represent articulated objects without considering the integrity of the articulated parts, which is a generic property shared by all articulated objects.
In this work, we utilise this property and leverage spatially continuous neural implicit representation to model the motion of the monolithic articulated parts.

\begin{figure}[h]
    \centering 
    \vspace{-6mm}
    \includegraphics[trim={45, 150, 60, 140}, clip, scale=0.45]{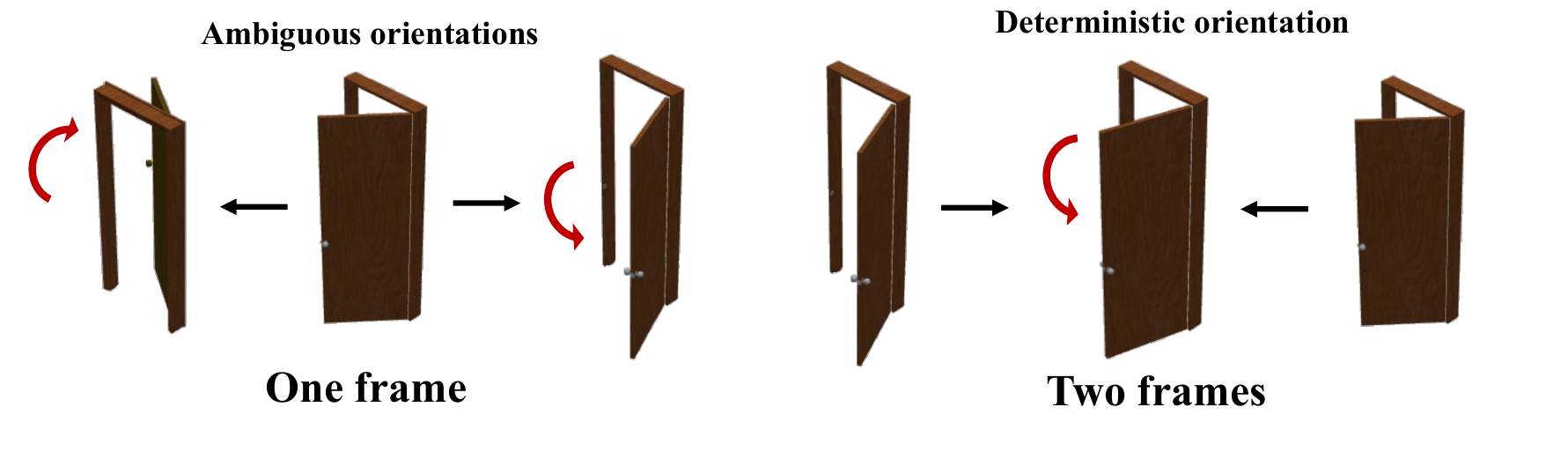}
    \caption{
    \textbf{Two point cloud frames are required} for learning articulated part motions, as one frame may indicate ambiguous motions (\emph{e.g.,} clockwise and anti-clockwise orientations).
    }
    \vspace{-4mm}
    \label{fig:two_part}
\end{figure}

\section{Problem Formulation}



Learning the motion of the articulated part on an object requires at least two frames of that object under different poses (\emph{e.g.,} different door opening degree).
That is because using one frame as the observation would have an ambiguity problem, take Figure~\ref{fig:two_part} as an example, given an observation of a door, we cannot distinguish whether the revolute direction is clockwise or anti-clockwise.

In this study, 
each object in training set provides two point cloud $I_1, I_2 \in \mathbb{R}^{N\times{3}}$ under different part poses. The model maps part motion to corresponding part pose scalar values $\phi_1, \phi_2 \in \mathbb{R}$ representing the degree of articulation, and can 1) generate new point cloud $I_3$ given a new part pose scalar $\phi_3 \in \mathbb{R}$.
2) can few-shot generalise to novel object categories.

%


\section{Method}
\label{sec:method}

\begin{figure*}[h]
    \centering
    \vspace{-3mm}
    \includegraphics[trim={5, 0, 15, 20}, clip, width=\linewidth, scale=1.15]{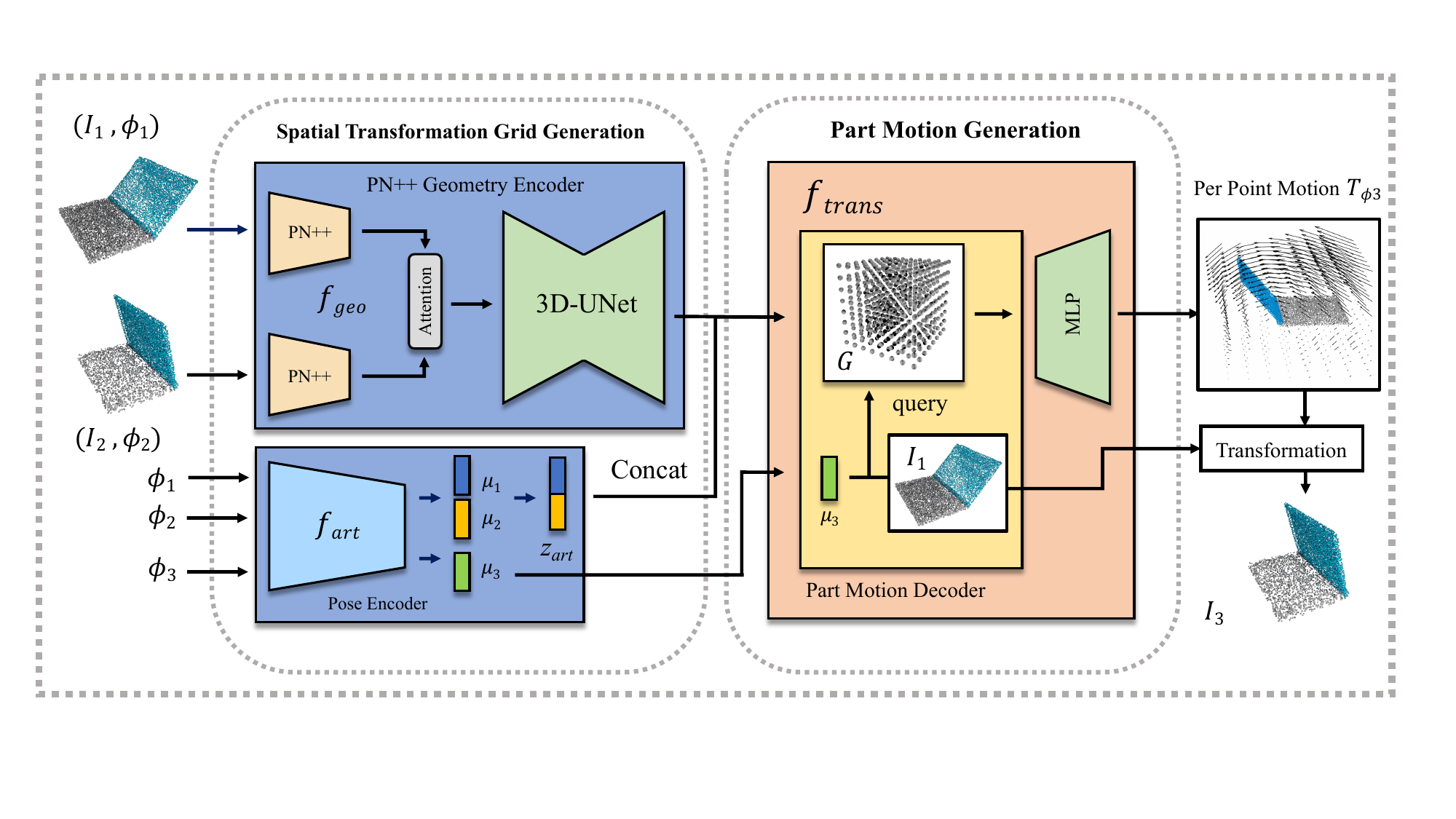}
    \vspace{-10mm}
    \caption{
    \textbf{Our proposed framework} receives two point clouds $I_1$ and $I_2$ from the same articulated object under two different part poses $\phi_1$ and $\phi_2$. Then generate the object point cloud $I_3$ with a new part pose $\phi_3$. It aggregates the geometric information of $I_1$ and $I_2$, and the pose information of $\phi_1$ and $\phi_2$ into a spatially continuous Transformation Grid. During inferencing, conditioned on the new part pose $\phi_3$, it decodes the transformation of each point by querying each point in the Grid to generate the input object with the novel pose.
    }
    \label{fig:pipeline}
\end{figure*}

As shown in Figure~\ref{fig:pipeline}, our proposed framework is mainly composed of two procedures, \textbf{Spatial Transformation Grid Generation} (Left) and \textbf{Part Motion Generation} (Right). 

\textbf{Spatial Transformation Grid Generation}:
As is described in \ref{sec:intro}. The distribution of the movements of points on the articulated part possesses spatial continuity over the 3D space.
In this section, our framework receives a pair of articulated object point clouds with their corresponding part poses $((I_1,\ \phi_1), (I_2,\ \phi_2))$, as well as a new part pose $\phi_3$ as input. Then output a Spatial Transformation Feature Grid $G$ to extract such spatial continuous features representing the part motions of articulated objects.

\textbf{Part Motion Generation}: 
In order to generate the object under pose $\phi_3$, we decode the transformation matrices from $\phi_1$ to $\phi_3$ of each point from the Spatial Transformation Feature Grid $G$.
Firstly, with respect to a novel part pose $\phi_3$, our framework retrieves each point $p$ in $I_1$'s transformation representation $\psi_{\phi_3} \in \mathbb{R}^{N \times d_{\psi}}$ in Spatial Transformation Grid $G$ using trilinear interpolation.
Then we decode each point's transformation representation into a transformation matrix $t_p$, and thus all the points' transformation matrix $t_p$ compose the whole transformation matrix $T_{\phi_3}$ for the whole point cloud $I_1$. 
Finally, we apply the transformation matrices $T_{\phi_3}$ to $I_1$ and get the point cloud $\hat{I_3}$ under the articulated part pose $\phi_3$. 
%
%
%
In the following sections, we show details of our proposed framework.


\vspace{-1mm}
\subsection{Spatial Transformation Grid Generation}

In this procedure, we generate the Spatial Transformation Feature Grid $G$ to extract the spatial distribution of joint motion features.

As mentioned in Section \ref{sec:intro}, the point motions on the articulated part surface  make up a continuous and smooth distribution with respect to point positions. 
Therefore, spatially continuous neural implicit representations are suitable for the representations of point motions.

We build such a 3D grid with $K \times K \times K$ points uniformly distributed in space ($K=32$), each point having implicit features representing both the object geometries and part motions.

To empower the learned Grid with both object geometries and part motions, the \textbf{Spatial Transformation Grid Generation} procedure consists of two submodules: 
1) \textit{Geometry Encoder} $f_{geo}$ that takes two point clouds under different part poses (which is $I_1$ and $I_2$) as input and outputs an implicit feature grid $G_{geo}$; 
2) \textit{Pose Encoder} $f_{art}$ that takes part poses $\phi_1$, $\phi_2$, $\phi_3$ respectively as input and outputs their respective features $\mu_1, \mu_2, \mu_3 \in \mathbb{R}^{d_{\mu}}$, and then concatenates $\mu_1, \mu_2$ into $z_{art}$ while passing down $\mu_3$ for further use. 
Finally, we concatenate $z_{art}$ to each grid feature of $G_{geo}$ to form Transformation Feature Grid $G$:
$$G_{geo} = f_{geo}(I_1,\ I_2),
z_{art} = f_{art}(\phi_1,\ \phi_2),
G = [G_{geo},\ z_{art}]$$

\vspace{-1mm}
\subsubsection{Geometry Encoder}


Inspired by Ditto~\cite{jiang2022ditto}, to extract geometric information of the two input point clouds, we first use PointNet++~\cite{qi2017pointnetplusplus} encoders to encode $I_1$ and $I_2$, and extract sub-sampled point cloud features $h_1, h_2 \in \mathbb{R}^{N' \times d}$, where $N'$ denotes the point number after the sub-sampling procedure of PointNet++, and we use $N' = 128$ in our work.

To aggregate the features of the two input point clouds, we employ an attention module between sub-sampled point features $h_1, h_2$ into an aggregated feature $h$:
$$ h = [h_1, softmax(\frac{h_1h_2^T}{\sqrt{d}})h_2] $$
Then, we feed the aggregated feature $h$ into a 3D-UNets~\cite{cciccek20163d}and generate 3D geometric implicit feature grid $G_{geo}$ representing geometric information of the two input point clouds with $K \times K \times K$ uniformly distributed points.


\subsubsection{Pose Encoder}
We use Multi-Layer Perceptrons (MLP) to separately encode part poses $\phi_1,\ \phi_2,\ \phi_3$ into articulation features $\mu_1,\ \mu_2,\ \mu_3\ \in \mathbb{R}^{N \times d_{art}}$, and concatenate $\mu_1$ and $\mu_2$ to form 
$ z_{art} = [\mu_1, \mu_2] $.

We again concatenate $z_{art}$ with each point feature in $G_{geo}$ to form Spatial Transformation Feature Grid $G$, containing spatially continuous implicit features about both the geometric information and the pose information of the target object in the space.


\vspace{-1mm}
\subsection{Part Motion Generation}

During the above \textbf{Spatial Transformation Grid Generation} procedure, we have generated the Spatial Transformation Feature Grid $G$. 
In this \textbf{Part Motion Generation} procedure, we use $G$ to generate spatially continuously distributed point motions from $I_1$ to the target $I_3$.


Firstly, from $G$ which is composed of $K \times K \times K$ points uniformly distributed in the space with their corresponding features, we query the feature $f_p$ under $\mu_3$ of each point $p$ on the articulated part using trilinear interpolation. 

Then, we employ a motion decoder $f_{trans}$ (composed of an MLP network) to decode the transformation matrix $t_p$ from $\phi_1$ to $\phi_3$ of each point $p$ on the articulated part. Taking pose feature $\mu_3$ as conditions, our decoder obtain the corresponding $t_p$ and conduct elemental-wise production to generate the point cloud prediction $\hat{I_3}$ under the part pose $\phi_3$. 
$$\psi_{\phi_3} = Query(I_1, G), 
T_{\phi_3} = f_{trans}(\psi_{\phi_3}), 
\hat{I_3} = T_{\phi_3} \cdot I_1  $$
In this way, for those points on the articulated parts, their motions could be generated smoothly from the spatially continuous distribution, keeping the part as a whole after the motion, while maintaining the geometric details of them.

\subsection{Training and Loss}

\textbf{Data collection.} 
To generate diverse data for training, we randomly sample articulated part poses $\phi_1$, $\phi_2$ and $\phi_3$ and then generate point cloud observation $I_1$, $I_2$ and $I_3$ corresponding to each part poses. Ascribing to the ability to get point could in simulator with arbitrary part poses, we can generate diverse $((I_1, \phi_1), (I_2, \phi_2), \phi_3)$ for training.

\vspace{1mm}
\textbf{Loss function.} 
We use Earth Mover's Distance (EMD)~\cite{rubner2000earth} as the loss function. 
EMD is utilised to estimate the distance between two distributions. We can calculate the EMD between two point clouds by calculating the minimum amount of point movements needed to change the generated object point cloud into the target. In our work, with the input data $((I_1, \phi_1), (I_2, \phi_2), \phi_3)$, the EMD is computed between the ground truth point cloud $I_3$ of the articulated object with the part pose $\phi_3$, and our prediction $\hat{I_3}$.

We set up a loss optimising whole point cloud and increase the weight of loss on movable part to facilitate neat part formulation with smooth surfaces and fewer outliers.
$$Loss = EMD(I_3, \hat{I_3})$$

\section{Experiments}
\label{sec:experiment}

We conduct our experiments using the large-scale PartNet-Mobility~\cite{chang2015shapenet, Mo_2019_CVPR, Xiang_2020_SAPIEN} dataset of 3D articulated objects, covering over 7 object categories. We evaluate the performance of our method in several tasks including: 1) the articulated object generation for unseen objects in training categories, 
2) few-shot articulated object generation for novel object categories, 
and 3) the interpolation and extrapolation of the spatial continuous NIR. 
Quantitative and qualitative results compared to several baselines and an ablated version demonstrate our method's superiority over other methods. 




\vspace{-5mm}
\subsection{Baselines and Metrics}

We evaluate and compare our approach with the following two baselines and one ablation:

\textbf{A-SDF}~\cite{mu2021sdf} represents objects with a shape code and an articulation code. Given an object, it first infers the shape and articulation codes and then generates the shape at unseen angles by keeping the shape code unchanged and changing the articulation code.

\textbf{Ditto}~\cite{jiang2022ditto} also takes two point clouds as input to learn the structure of an articulated object. It directly predicts the occupancy, the segmentation, and the joint configuration to build a digital twin. The original paper demonstrates the point cloud reconstruct ability, we modify it to take a new part pose as input and then generate the corresponding object.

\textbf{Ours w/o NIR} is an ablated version of our method that directly predicts the transformation matrix for each point to generate the new point cloud without applying spatially continuous NIR as a middle step. We conduct this ablation version to demonstrate the effectiveness of our design using Spatial Transformation Feature Grid $G$.

To evaluate the generated objects and their similarity with the ground-truth objects, we apply the Earth Mover's Distance (EMD)~\cite{rubner2000earth} as the evaluation metric.



\vspace{-5mm}
\subsection{Evaluation on Unseen Objects in Training Categories}

\begin{figure*}[h]
    \centering  
    \includegraphics[trim={55, 20, 160, 0}, clip, scale=0.45]{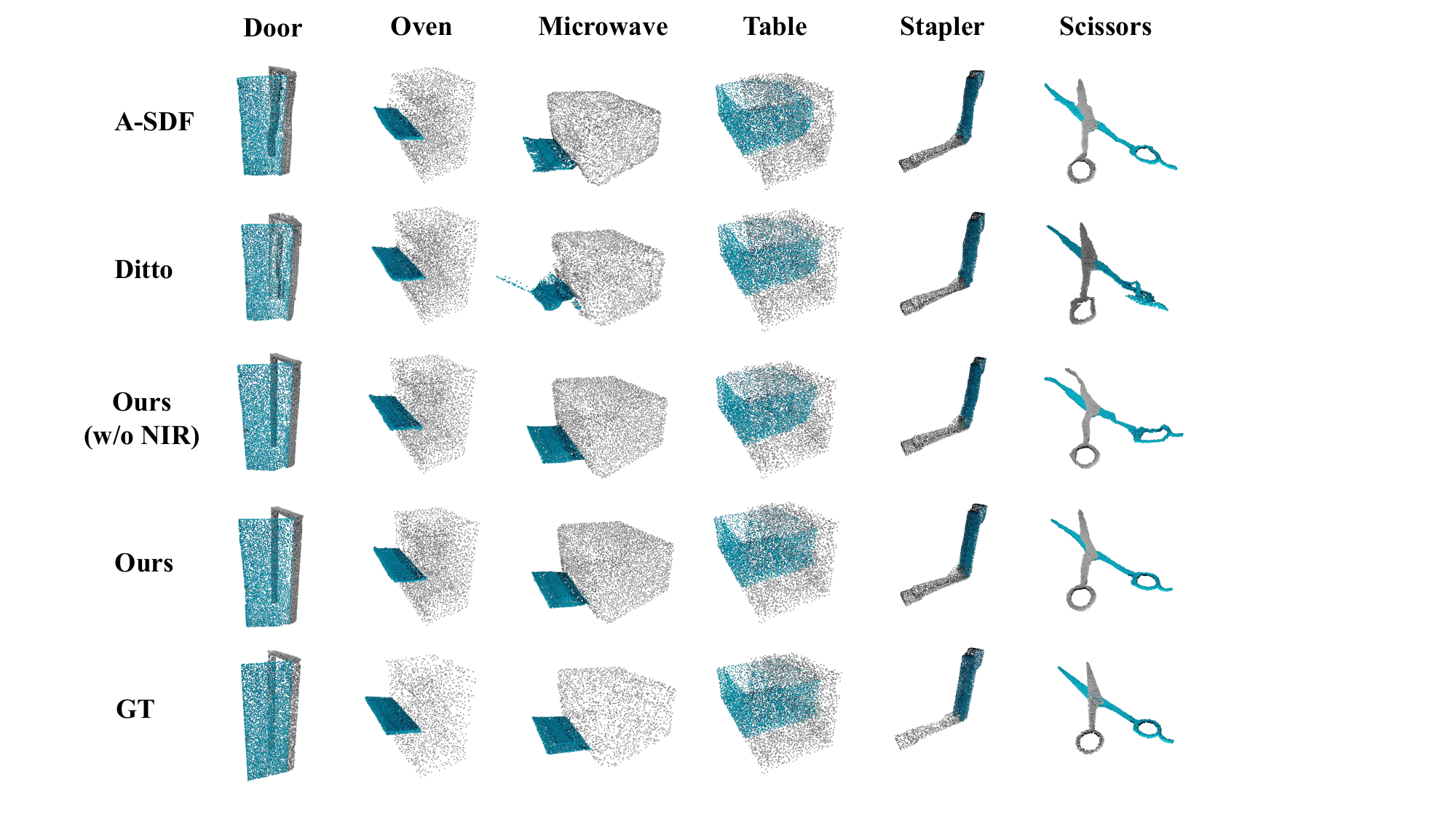}
    \vspace{5mm}
    \caption{
    \textbf{Visualisation of generated objects in training categories} shows our method reserves the most detailed geometries of both articulated parts and object bases.
    For example, our model predicts the straightest door frame and the smoothest microwave door surface.
    }
    \vspace{-2mm}
    \label{fig:single_cat}
\end{figure*}

\vspace{-2mm}
\begin{table*}[htbp]
  \centering
  \resizebox{\textwidth}{!}{
    \begin{tabular}{@{}lcccccccccc@{}}
    \toprule
    Method              & Laptop & Stapler & Door & Scissors & Oven & Refrigerator & Microwave & Table &  \\ \midrule \midrule
    A-SDF & 1.6923 & 3.9335 & 3.2459 & 1.9307 & 1.3983 & 2.2532 & 3.7570 & 1.8467 &  \\ \midrule
    Ditto    & 1.6195 & 3.1161 &        2.9811 & 2.1619 & 1.3401 & 1.9863 & 4.8210 & 1.4010 &  \\ \midrule
    Ours w/o NIR & 1.6080 & 3.3369 & 2.5863 &  2.0628 &  1.1294  & 2.1539 & 1.9281 & 1.5189 & \\ \midrule
    Ours & \textbf{1.4420} & \textbf{3.0850} & \textbf{2.2808} & \textbf{1.8025}  &  \textbf{1.1134}  & \textbf{1.6431}  & \textbf{1.8088} & \textbf{1.3315} &     \\ \bottomrule
    \end{tabular}
    }
  \vspace{4mm}
  \caption{\textbf{Earth Mover’s Distance (EMD) on object generation in training categories.} 
  }
  \label{tab_quan_train}
  \vspace{-2mm}
\end{table*}

In this task, given an articulated object in the training category with two point clouds and the corresponding part poses, we generate its point cloud with novel part poses. 

The quantitative results in Table~\ref{tab_quan_train} demonstrate that our proposed framework outperforms all other methods in all categories with lower EMD, which means that our generated articulated objects are the closest to the ground-truth shapes. The qualitative results in Figure~\ref{fig:single_cat} also show that our generated objects reserve the most detailed geometry. In comparison, the performance of both Ditto and A-SDF is worse, for example, they both fail to predict the door frame straightly, and fail to predict the microwave door surface smoothly. 

The main reason for the difference is, A-SDF and Ditto directly decode the whole point cloud into latent space, while ours takes the integrity of parts into consideration by querying the motion of each point in the original point cloud. This one-to-one mapping from the original shape to the generated shape best preserves geometric features of the original shape.

\vspace{-1mm}
\subsection{Evaluation on Novel Categories} 

In this task, we use the pretrained model in one category and finetune the model in a novel category using only a few objects for a few epochs. Specifically, we use 8 objects in the novel category, and the finetuning time is one-twentieth of the training time from scratch.
It is worth mentioning that the directions of the articulated part axes in the training set and finetuning set are different in these experiments (\emph{i.e.,} we train on the up-down opening ovens and finetune on the left-right opening refrigerators.)
This task aims to demonstrate that learning the part motions of articulated objects makes the model easier to adjust to a novel kind of articulated object, as it is the shared property of all articulated objects. 

\begin{wraptable}{r}{6cm}
  \centering
  \vspace{-2mm}
    \begin{tabularx}{6.12cm}{@{}l>{\centering\arraybackslash}m{0.95cm}>{\centering\arraybackslash}m{0.95cm}>{\centering\arraybackslash}m{0.95cm}>{\centering\arraybackslash}m{0.95cm}@{}}
    \toprule
    Method   & Oven-Refri & Refri-Oven & Door-Laptop & \\ \midrule
    \midrule
    A-SDF    & 37.7618 & 2.0164 & 2.2532 & \\ \midrule
    Ditto    & 3.9443 & 2.1832 & 2.3997 & \\ \midrule
    Ours w/o NIR & 15.5203 & 1.6832 & 2.3547 & \\ \midrule
    Ours & \textbf{2.5794} & \textbf{1.5440} & \textbf{2.0873} &   \\ \bottomrule
    \end{tabularx}
  \vspace{4mm}
  \caption{\textbf{Earth Mover’s Distance (EMD) on object generation in novel categories.} 
  }
  \vspace{-6mm}
  \label{tab_quan_novel}
\end{wraptable}

\begin{wrapfigure}{l}{6cm}
    \centering 
    \vspace{-2mm}
    \includegraphics[trim={260, 10, 300, 42}, clip, scale=0.42]{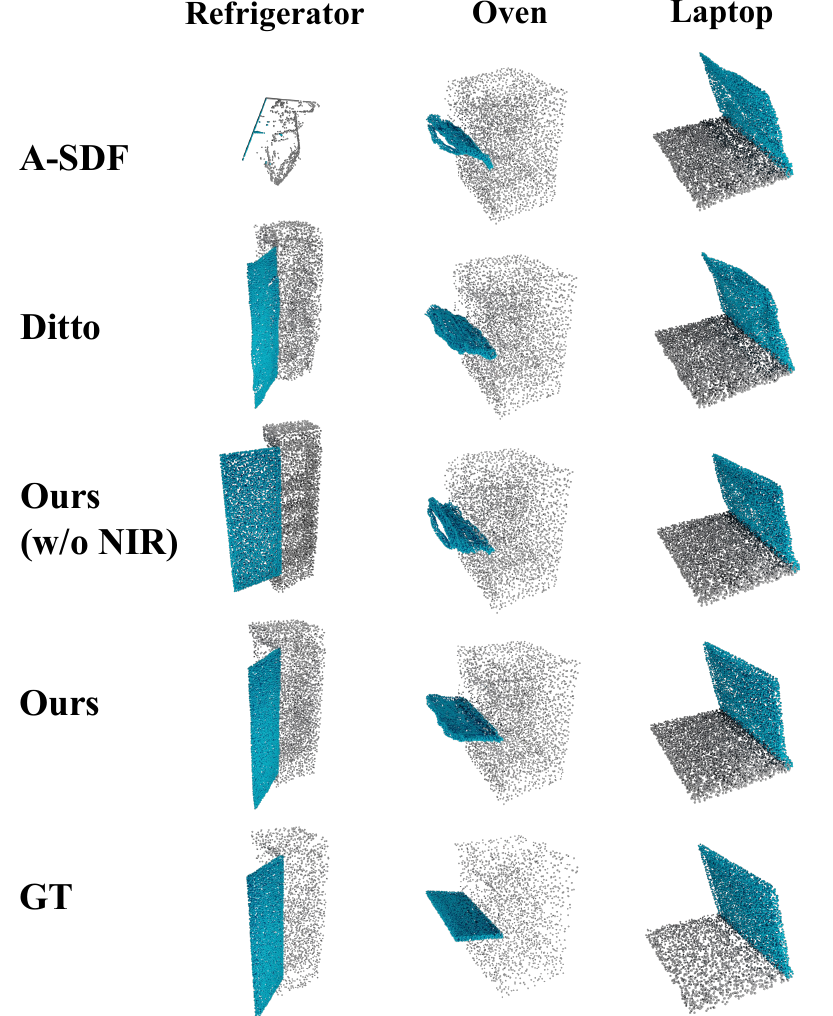}
    \vspace{0.5mm}
    \caption{
    \textbf{Visualisation of generated objects in novel categories} shows our method maintains geometric consistency. 
    }
    \vspace{-1.2cm}
    \label{fig:finetune}
\end{wrapfigure}

The quantitative results in Table~\ref{tab_quan_novel} show that our method achieves significantly better results with lower EMD compared to all the baselines, especially in the Oven-Refrigerator block. The visualisation results of Figure~\ref{fig:finetune} also show that our method present the most accurate part poses and the most precise part geometry after a short-period finetuning.

Failures of A-SDF possibly come from that, the representations learned by A-SDF are limited to the trained articulated object category and are hard to adjust to novel shapes and articulations. 

We have also conducted experiments using the widely-used metric Pose Angle Error (PAE) and Chamfer Distance (CD), with results shown in Table~\ref{tab_cd_pae}.

\begin{wraptable}{r}{6.5cm}
  \centering
  \vspace{0mm}
      {
  \begin{tabular}{lcccc}
    \toprule
    Metric & A-SDF & Ditto & Ours \\ 
    \midrule
    CD $\downarrow$ & 2.213 & 2.019 & \textbf{1.782}\\ 
    PAE (degree) $\downarrow$ & 6.457 & 6.212 & \textbf{4.767}\\
    \bottomrule
  \end{tabular}
  }
  \vspace{5mm}
  \caption{\textbf{Evaluations on CD and PAE.}
  \vspace{-2mm}
  }
  \label{tab_cd_pae}
\end{wraptable}

Our superior performance in novel categories against Ditto mainly comes from the use of transformation matrix to represent part motion. 
Intuitively, a transformation matrix could represent any kind of motion in 3D space and is spatial continuous for points on the motion part. As a result, it has the potential to few-shot generalise to any kind of part motion no matter its displacement.
\vspace{-1mm}

\subsection{Ablation Studies and Analysis}
\vspace{1mm}

We compare our method with the ablated version without Neural Implicit Representations (\textbf{Ours w/o NIR}). Results in Table~\ref{tab_quan_train} and Table~\ref{tab_quan_novel} show that NIR helps the generated point cloud to be closer to the ground-truth target, representing by the lower EMD between the generated objects and the ground-truth objects. 
From the visualisation in Figure~\ref{fig:single_cat} and Figure~\ref{fig:finetune}, we can observe that the point clouds generated with NIR have more accurate part pose and smoother part surface. Those results demonstrate that by using Spatially Continuous Neural Implicit Representation to model the part motion, our framework gets a better distribution for motion representations in the 3D space.


\vspace{-0.3cm}

\subsection{Analysis of Transformation on Grid Points}
\label{sec:tranformation}
Figure~\ref{fig:inter_NIR} visualises the transformation grid of a refrigerator instance (in the first row) and an oven instance (in the second row). 
The figures on the left are displayed in 3D while the right ones are displayed in 2D. Note that for better visualisation and understanding, on the right, we represent the refrigerator in the top-down view, and represent the oven in the side view. 
The arrows forging circles centreing the ground-truth joint show that our model successfully projects the part motion to euclidean space. 

\subsection{Interpolation and Extrapolation} 
\vspace{-1mm}

\begin{figure*}[t]
    \centering  
    \includegraphics[trim={550, 440, 550, 350}, clip, scale=0.45]{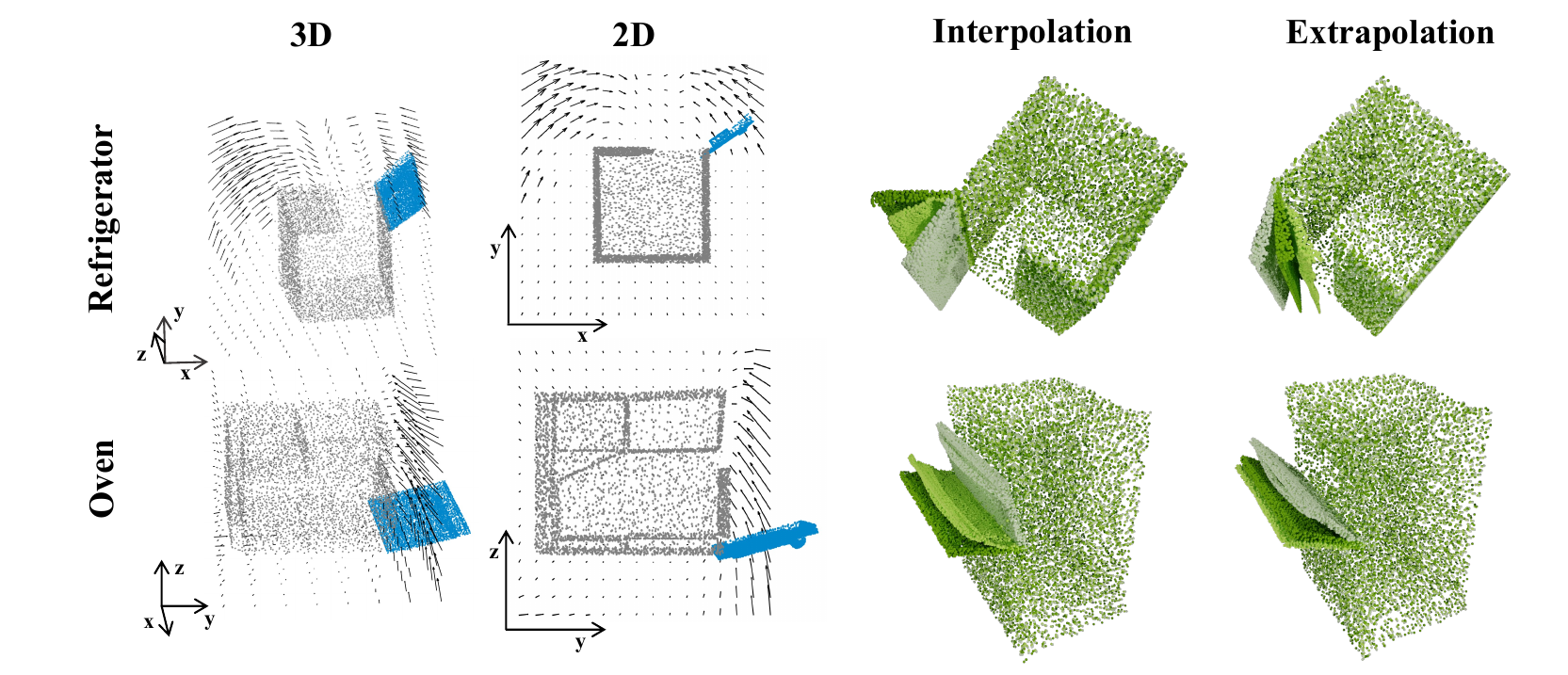}
    \vspace{4mm}
    \caption{
    \textbf{Visualisation of the transformation on grid points (left), and results of interpolation and extrapolation (right).}
    }
    \vspace{-6mm}
    \label{fig:inter_NIR}
\end{figure*}

\begin{table}[t]
  \centering
  \vspace{4mm}
    \begin{tabular}{@{}ccccc|cccc@{}}
    \toprule
    Method   & Fridge & Oven & Door & Table & Fridge & Oven & Door & Table \\
    \midrule \midrule
    A-SDF    & 4.1248 & 1.6185 & 2.8883 & 8.3870 & 4.2671 & 2.5514 & 5.3937 & 8.0931 \\
    \midrule
    Ditto    & 3.2421 & 1.2861 & 2.5974 & 9.8180 & 3.1738 & 1.9405 & 4.4676 & 8.5025 \\
    \midrule
    Ours     & \textbf{2.1256} & \textbf{1.2364} & \textbf{2.1330} & \textbf{8.1688} & \textbf{2.8376} & \textbf{1.7669} & \textbf{3.5298} & \textbf{7.0804} \\ \bottomrule
    \end{tabular}
  \vspace{4mm}
  \caption{
    \textbf{EMD on interpolation (Left) and extrapolation (Right) results.}
    }
  \label{tab_inter_extrapolation}
  \vspace{-5mm}
\end{table}

Interpolation and extrapolation between shapes is a key ability for 3D object representations which reveals the distribution of articulation part poses. 

In this task, given two shapes of the same object, we generate the object with novel articulation degrees in between or beyond. 
In Table~\ref{tab_inter_extrapolation}, quantitative results show that our method outperforms A-SDF and Ditto in both interpolation and extrapolation tasks.
In Figure~\ref{fig:inter_NIR}, we represent the input parts with dark and light green, and the generated part with medium green. The results demonstrate our representation of part motion is continuous and dense.







\begin{wrapfigure}{r}{6cm}
    \centering 
    \vspace{-10mm}
    \includegraphics[trim={0, 0, 0, -10}, clip, scale=0.3]{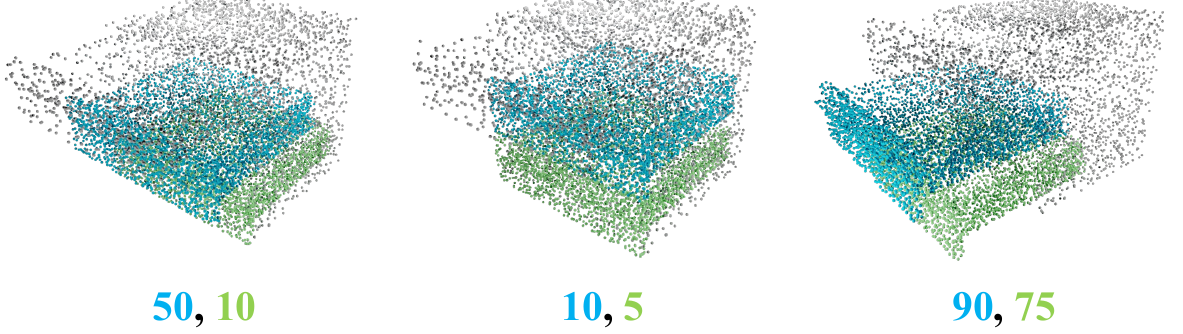}
    \vspace{1mm}
    \caption{
    \textbf{Multi-part object generation.}
    }
    \vspace{-10mm}
    \label{fig:multi_part}
\end{wrapfigure}

\vspace{-3mm}
\subsection{Multi-part Generation}
Our method can easily extend to objects with multiple  parts by changing the input part angle to a vector of part angles, shown in Figure~\ref{fig:multi_part}.
\vspace{-1mm}
\section{Conclusion}
\vspace{-2mm}
\label{sec:conclusion}

In this paper, 
we propose a novel framework for modelling and generating articulated objects. 
To model the continuous articulations and motions smoothly, we adopt neural implicit representations (NIR) to predict the transformations of moving part points of the object. Experiments on different representative tasks 
demonstrate that our proposed framework outperforms other methods both quantitatively and qualitatively.

\vspace{-3mm}
\section{Acknowledgment}
\vspace{-3mm}

This work was supported by National Natural Science Foundation of China (No. 62136001).

\bibliography{egbib}
\end{document}